\documentclass[11pt]{article}

\usepackage[final]{acl}
 \usepackage{amsmath}

\usepackage{times}
\usepackage{latexsym}

\usepackage[T1]{fontenc}

\usepackage[utf8]{inputenc}

\usepackage{microtype}

\usepackage{inconsolata}

\usepackage{graphicx}

\title{Lexicon-Enriched Graph Modeling for Arabic Document Readability Prediction}

\author{
  Passant Elchafei\thanks{Equal contribution.} \\
  Ulm University, Germany \\
  \texttt{passant.elchafei@uni-ulm.de}
  \And
  Mayar Osama\footnotemark[1] \\
  German University in Cairo, Egypt \\
  \texttt{mayar.osama@guc.edu.eg}
  \AND
  Mohamed Rageh \\
  German University in Cairo, Egypt \\
  \texttt{mohamad.rageh@student.guc.edu.eg}
  \And
  Mervat Abuelkheir \\
  German University in Cairo, Egypt \\
  \texttt{mervat.abuelkheir@guc.edu.eg}
}

\begin{document}
\maketitle
\begin{abstract}
We present a graph-based approach enriched with lexicons to predict document-level readability in Arabic, developed as part of the Constrained Track of the BAREC Shared Task 2025. Our system models each document as a sentence-level graph, where nodes represent sentences and lemmas, and edges capture linguistic relationships such as lexical co-occurrence and class membership. Sentence nodes are enriched with features from the SAMER lexicon as well as contextual embeddings from the Arabic transformer model. The graph neural network (GNN) and transformer sentence encoder are trained as two independent branches, and their predictions are combined via late fusion at inference. For document-level prediction, sentence-level outputs are aggregated using max pooling to reflect the most difficult sentence. Experimental results show that this hybrid method outperforms standalone GNN or transformer branches across multiple readability metrics. Overall, the findings highlight that fusion offers advantages at the document level, but the GNN-only approach remains stronger for precise prediction of sentence-level readability.
\end{abstract}

\section{Introduction}

Accurately assessing the readability of Arabic documents is essential for educational technologies, language learning platforms, and adaptive content delivery systems. The task poses significant linguistic challenges due to the diglossic nature of Arabic, rich morphology, and the scarcity of large-scale annotated corpora~\cite{imperial-kochmar-2023-automatic}. The BAREC Shared Task 2025 \cite{elmadani-etal-2025-barec-shared-task} addresses this by providing a fine-grained classification benchmark: assigning one of 19 readability levels to Arabic texts at both the sentence and document level.

Previous work on Arabic NLP has applied deep contextual models such as BERT variants for various classification tasks, including readability prediction ~\cite{article, antoun-etal-2020-arabert}. Although effective, these approaches typically operate only on text sequences and often overlook explicit structural and lexical relationships that can influence readability. In contrast, graph-based methods make it possible to encode document-level structure and linguistic relationships directly ~\cite{SUN2023110428}. In this work, we explicitly incorporate such relationships by leveraging the SAMER lexicon for lexical difficulty features and constructing a heterogeneous sentence-lemma graph with multiple edge types (e.g., HAS\_LEMMA, OCCUR\_WITH, IN\_CLASS, IN\_DOMAIN). This allows our model to combine the strengths of contextual embeddings with explicit lexical and structural graph modeling, which we show experimentally to improve both sentence-level and document-level readability prediction.

We propose a hybrid approach that represents each document as a graph, where nodes correspond to sentences and lemmas, and edges represent linguistic relationships such as \texttt{HAS\_LEMMA}, \texttt{OCCUR\_WITH}, and \texttt{IN\_CLASS}. Each sentence node is enriched with difficulty signals from the SAMER lexicon~\cite{al-khalil-etal-2020-large} and contextual sentence embeddings from the \texttt{readability-arabertv2-d3tok-CE} model, a fine-tuned variant of AraBERTv2 optimized for Arabic readability classification~\cite{antoun-etal-2020-arabert}.

To integrate both modalities, we train the GNN (graph modality) and the transformer (text modality) independently and use late fusion to merge their readability predictions at the end of inference. This approach combines the strengths of structured lexical-graph features and contextual text embeddings, without mixing intermediate features. Document-level labels are then obtained by pooling the sentence-level predictions, using max-pooling to reflect the most difficult sentence.

Our experiments demonstrate that this lexicon-enriched, confidence-aware, graph-based approach significantly improves prediction performance over individual branches. The results emphasize the importance of combining structured lexical knowledge with neural contextualization and fusion to better capture Arabic document readability.

\section{Related Work}
Automatic readability assessment has become a key area in NLP due to its applications in education, text simplification, and adaptive content delivery. In English, early studies relied on surface-level features such as sentence length and word frequency, followed by statistical models and, more recently, neural methods that capture semantic and discourse-level information \cite{imperial-kochmar-2023-automatic}.

For Arabic, early research was constrained by resource scarcity and linguistic complexity. One of the first efforts was the AARI index \cite{article}, which used handcrafted lexical and syntactic features derived from academic curricula. Later, the SAMER Lexicon \cite{al-khalil-etal-2020-large} introduced a large-scale graded vocabulary resource. Subsequently, it was showcased in a word-level readability visualization system designed for assisted text simplification \cite{hazim-etal-2022-arabic}. More recently, the SAMER Corpus \cite{alhafni-etal-2024-samer} provided the first manually annotated Arabic parallel dataset for text simplification targeting school-aged learners. These resources provided the foundation for subsequent work.

In recent years, several datasets have advanced Arabic readability modeling. The BAREC corpus \cite{elmadani-etal-2025-large} provides a large-scale benchmark with 19 readability levels at both the sentence and document level, while the DARES dataset \cite{el-haj-etal-2024-dares} focuses on Saudi school textbooks. Complementary approaches, such as AraEyebility \cite{computation13050108}, integrate eye-tracking signals to connect human cognitive processing with readability prediction. In addition, \cite{liberato-etal-2024-strategies} explored strategies for Arabic readability modeling, highlighting the need to combine lexical resources with modern learning-based approaches. A survey by \cite{CAVALLISFORZA201838} provides an overview of the challenges and future directions for Arabic readability assessment.

Overall, most Arabic readability models have focused on surface features or contextual embeddings in isolation, with limited integration of structured lexical knowledge. To our knowledge, no prior work has combined lexicon-enrichment with graph-based modeling for Arabic document readability. Our work addresses this gap by integrating the SAMER Lexicon into a heterogeneous sentence-lemma graph, capturing both vocabulary difficulty and structural relations to improve fine-grained readability prediction.

\section{System Overview}
The purpose of our approach is to capture the linguistic characteristics and the relationships between the features of two datasets: BAREC \cite{elmadani-etal-2025-large} and SAMER \cite{al-khalil-etal-2020-large}. The BAREC dataset consists of \textbf{sentences} annotated with their corresponding \textbf{readability levels}. The SAMER dataset consists of \textbf{lemmas}, each associated with an average \textbf{readability level} across different dialects, along with additional features such as frequency of occurrence and part-of-speech (POS) tags for each \textit{(lemma, readability level)} pair.

We integrate the two datasets by extracting lemmas from the sentences while preserving their POS tags and recording the count of diacritics. The extraction of lemmas was performed using the CAMeL Tools Morphology Analyzer \cite{cameltools}. Each extracted lemma is then matched against the SAMER lexicon to enrich it with statistical attributes such as average readability, frequency, and POS. This alignment ensures that the SAMER lexicon contributes directly to the graph as node features rather than as isolated entries.\\

The combined data is reformulated into a heterogeneous graph $\mathcal{G} = (\mathcal{V}, \mathcal{E})$ consisting of multiple node and edge types. The node set $\mathcal{V}$ includes:

\begin{itemize}
    \item \textbf{Sentences}: Represented by 768-dimensional embeddings obtained from the CAMeL-Lab Arabic readability model (readability-arabertv2-d3tok-CE), augmented with linguistic features.
    \item \textbf{Lemmas}: Characterized by statistical attributes such as average readability and frequency.
    \item \textbf{Classes}: Educational difficulty levels, one-hot encoded as Foundational, Advanced, or Specialized.
    \item \textbf{Domains}: Subject domains encoded as Arts \& Humanities, STEM, or Social Sciences.
\end{itemize}
The main objective of our approach is to leverage both classical linguistic features from the two datasets and deep Graph Neural Networks (GNNs) to capture hidden patterns within the data.
\begin{figure*}
  \centering
    \includegraphics[width=0.9\textwidth,height=4.5cm]{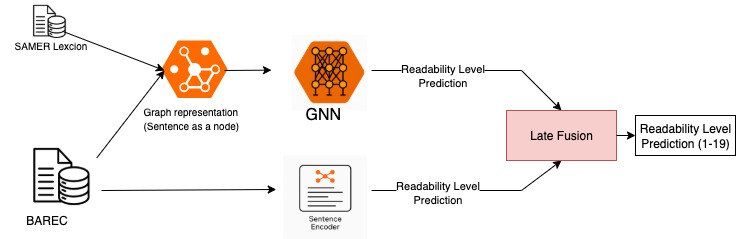}
  \caption{Overview of our proposed hybrid architecture for Arabic readability prediction. Sentence-level graphs are constructed using lexical relations from the SAMER lexicon and structural information from BAREC data. The GNN branch processes the graph to produce a softmax probability distribution over 19 readability levels, while the sentence encoder branch generates parallel probabilities from contextual embeddings. Inference uses late fusion, where both probability vectors are combined at the prediction stage using a tunable weight, yielding the final readability level for each sentence or aggregated document.}
  \label{fig:framework}
\end{figure*}
The edge set $\mathcal{E}$ contains the following directed relations:
\begin{itemize}
    \item \texttt{sentence $\rightarrow$ lemma} (\textit{HAS\_LEMMA}): Indicates lexical composition.
    \item \texttt{lemma $\leftrightarrow$ lemma} (\textit{OCCUR\_WITH}): Represents lemma co-occurrence in context.
    \item \texttt{sentence $\rightarrow$ class} (\textit{IN\_CLASS}): Connects each sentence to its labeled difficulty class.
    \item \texttt{sentence $\rightarrow$ domain} (\textit{IN\_DOMAIN}): Links sentences to their broader academic domain.
\end{itemize}
Once the data is structured into the graph format, the first step is to apply input feature transformation, where we transform the node features into the model's hidden dimensions, for which we used a linear layer between the original dimensions to the target dimension to find optimal projection.

\[h_v^{(0)} = W_{\text{in}}^{(\tau)} x_v, \quad \text{for } v \in \mathcal{V}_\tau
\]

where $h_v^{(0)}$ is the initial hidden representation of node $v$ after projection, $W_{\text{in}}^{(\tau)}$ is the trainable weight matrix for input transformation for node type, $\mathcal{V}_\tau$ denotes nodes of type $\tau$, and $x_v$ is the raw feature vector.\\
The core of the model consists of a stack of SAGEConv \cite{hamilton2017inductive} hidden layers. Each is used to learn the graph embeddings over the heterogeneous graph. It uses neighbor sampling and aggregation. Each layer applies a learnable linear transformation to the combined features; this transformation allows the model to learn complex feature interaction while maintaining consistent dimensions across the layers.

The model consistes of 4 GNN layers, for which we use ReLU activation function $\sigma$ and layer normalization to avoid linearity and improve the gradient flow. 

{\small
\[
h_v^{(k)} = \sigma \left( \text{AGGREGATE}_{\text{type}} \left( \left\{ h_u^{(k-1)} : u \in \mathcal{N}_\text{type}(v) \right\} \right) \right)
\]
}
where $h_v^{(k)}$is the hidden representation of node $v$ at layer $k$, $h_u^{(k-1)}$ is the hidden representation of neighbor node $u$ from the previous layer, and $\mathcal{N}_\text{type}(v)$ denotes the set of neighboring nodes of $v$ connected via a specific edge type.
Additionally, we use a residual connection per layer. This preserves the features and provides more stable training, especially for the sentence nodes. 
\[
h_v^{(k)} \leftarrow \text{LayerNorm}\left(h_v^{(k)} + h_v^{(k-1)}\right)
\]
Finally, an MLP layer used for the classification.
\[
y_v = \text{MLP}(h_v^{(L)})
\]

\section{Experimental Results}
We conduct experiments on both sentence-level and document-level readability prediction tasks, as defined in the BAREC Shared Task 2025. For sentence-level classification, each sentence is represented as a node in the graph and labeled with one of 19 readability levels. For document-level prediction, we reuse the same model architecture and apply aggregation over sentence-level predictions. Specifically, we take the most difficult predicted sentence level (i.e., max pooling) as the document’s predicted readability level based on the intuition that the most complex sentence may determine the document’s comprehensibility floor.
\begin{table*}[t]
\centering
\begin{tabular}{|l|l|c|c|c|c|c|c|c|}
\hline
\textbf{Task Level} & \textbf{Model Variant} & \textbf{QWK} & \textbf{Acc} & \textbf{Acc +/-1} & \textbf{Dist} & \textbf{Acc 7} & \textbf{Acc 5} & \textbf{Acc 3} \\
\hline
Document-Level & GNN Only    & 75.6 & 40.0 & 83.0 & 0.8 & 60.0 & 60.0 & 90.0 \\
Document-Level & Late Fusion & \textbf{76.9} & \textbf{42.0} & 82.0 & 0.8 & 60.0 & 61.0 & 90.0 \\
Sentence-Level & GNN Only & \textbf{78.5} & \textbf{50.0} & 67.2 & 1.3 & 61.2 & 66.1 & 74.9 \\
Sentence-Level & Late Fusion & \textbf{78.5} & 41.4 & 65.9 &	1.4	& 55.4 & 62.6 & 72.7\\
\hline
\end{tabular}
\caption{Performance of the GNN-based model and Late Fusion on sentence-level and document-level readability prediction, evaluated with Quadratic Weighted Kappa (QWK), accuracy, accuracy within $\pm$1, distribution score, and accuracy at multiple granularity levels (7, 5, and 3).}
\label{tab:gnn-fusion-results}
\end{table*}

We evaluate two configurations:
\begin{itemize}
    \item \textbf{Late Fusion:} Combining weighted outputs from the GNN and transformer-based sentence encoder.
    \item \textbf{GNN Only:} Using the graph-based model without fusion.
\end{itemize}

The results in Table \ref{tab:gnn-fusion-results} show distinct trends between sentence-level and document-level tasks. For document-level prediction, Late Fusion outperforms the GNN-only baseline in both Quadratic Weighted Kappa (QWK; 76.9\% vs. 75.6\%) and exact accuracy (42.0\% vs. 40.0\%), while maintaining similar scores in the other metrics. QWK is a standard evaluation metric for ordinal classification that accounts for the degree of disagreement between predicted and true labels, making it particularly relevant for readability level prediction.

In contrast, for sentence-level prediction, the GNN-only model achieves substantially higher accuracy (50.0\% vs. 41.4\%) and better results in most metrics, despite both models having the same QWK (78.5\%). This indicates that, at the finer sentence granularity, the graph-based model alone is more effective, while the fusion approach may dilute some of the GNN’s discriminative power for exact classification.

Overall, the findings highlight that fusion offers advantages at the document level, but the GNN-only approach remains stronger for precise sentence-level readability prediction.
\section{Conclusion}
In this paper, we proposed a hybrid approach for Arabic document readability prediction by combining graph-based reasoning with contextual transformer-based modeling. Our architecture integrates lexical difficulty knowledge from the SAMER lexicon, sentence embeddings from a fine-tuned AraBERTv2 variant, and a structured graph representation of each document.

For sentence-level prediction, we demonstrated the benefits of lexicon-enriched heterogeneous graph modeling using a weighted GNN. For document-level prediction, we reuse the same graph setup and infer the document’s label by selecting the maximum difficulty among sentence-level predictions. This design aligns with the task’s objective of identifying the highest comprehension barrier within a document.

By applying late fusion between the GNN and transformer predictions, we achieved stronger performance across both levels. Our results highlight the complementary nature of structural and contextual signals and the promise of fusion-based systems for fine-grained Arabic readability tasks.

\bibliography{acl_latex}
\end{document}